# Decidable Reasoning in Terminological Knowledge Representation Systems

**Martin Buchheit**                                             BUCHHEIT@DFKI.UNI-SB.DE
*German Research Center for Artificial Intelligence (DFKI)*
*Stuhlsatzenhausweg 3, D-66123 Saarbrücken, Germany*

**Francesco M. Donini**                                         DONINI@ASSI.DIS.UNIROMA1.IT
**Andrea Schaerf**                                              ASCHAERF@ASSI.DIS.UNIROMA1.IT
*Dipartimento di Informatica e Sistemistica*
*Università di Roma "La Sapienza", Via Salaria 113, I-00198 Roma, Italy*

## Abstract

Terminological knowledge representation systems (TKRSs) are tools for designing and using knowledge bases that make use of terminological languages (or concept languages). We analyze from a theoretical point of view a TKRS whose capabilities go beyond the ones of presently available TKRSs. The new features studied, often required in practical applications, can be summarized in three main points. First, we consider a highly expressive terminological language, called $\mathcal{ALCNR}$, including general complements of concepts, number restrictions and role conjunction. Second, we allow to express inclusion statements between general concepts, and terminological cycles as a particular case. Third, we prove the decidability of a number of desirable TKRS-deduction services (like satisfiability, subsumption and instance checking) through a sound, complete and terminating calculus for reasoning in $\mathcal{ALCNR}$-knowledge bases. Our calculus extends the general technique of constraint systems. As a byproduct of the proof, we get also the result that inclusion statements in $\mathcal{ALCNR}$ can be simulated by terminological cycles, if descriptive semantics is adopted.

## 1. Introduction

A general characteristic of many proposed terminological knowledge representation systems (TKRSs) such as KRYPTON (Brachman, Pigman Gilbert, & Levesque, 1985), NIKL (Kaczmarek, Bates, & Robins, 1986), BACK (Quantz & Kindermann, 1990), LOOM (MacGregor & Bates, 1987), CLASSIC (Borgida, Brachman, McGuinness, & Alperin Resnick, 1989), KRIS (Baader & Hollunder, 1991), K-REP (Mays, Dionne, & Weida, 1991), and others (see Rich, editor, 1991; Woods & Schmolze, 1992), is that they are made up of two different components. Informally speaking, the first is a general schema concerning the classes of individuals to be represented, their general properties and mutual relationships, while the second is a (partial) instantiation of this schema, containing assertions relating either individuals to classes, or individuals to each other. This characteristic, which the mentioned proposals inherit from the seminal TKRS KL-ONE (Brachman & Schmolze, 1985), is shared also by several proposals of database models such as Abrial's (1974), CANDIDE (Beck, Gala, & Navathe, 1989), and TAXIS (Mylopoulos, Bernstein, & Wong, 1980).

Retrieving information in actual knowledge bases (KBs) built up using one of these systems is a deductive process involving both the schema (TBox) and its instantiation (ABox).

In fact, the TBox is not just a set of constraints on possible ABoxes, but contains intensional information about classes. This information is taken into account when answering queries to the KB.

During the realization and use of a KB, a TKRS should provide a mechanical solution for at least the following problems (from this point on, we use the word *concept* to refer to a class):

1. *KB-satisfiability*: are an ABox and a TBox consistent with each other? That is, does the KB admit a model? A positive answer is useful in the validation phase, while the negative answer can be used to make inferences in refutation-style. The latter will be precisely the approach taken in this paper.

2. *Concept Satisfiability*: given a KB and a concept $C$, does there exist at least one model of the KB assigning a non-empty extension to $C$? This is important not only to rule out meaningless concepts in the KB design phase, but also in processing the user's queries, to eliminate parts of a query which cannot contribute to the answer.

3. *Subsumption*: given a KB and two concepts $C$ and $D$, is $C$ more general than $D$ in any model of the KB? Subsumption detects implicit dependencies among the concepts in the KB.

4. *Instance Checking*: given a KB, an individual $a$ and a concept $C$, is $a$ an instance of $C$ in any model of the KB? Note that retrieving all individuals described by a given concept (a *query* in the database lexicon) can be formulated as a set of parallel instance checkings.

The above questions can be precisely characterized once the TKRS is given a semantics (see next section), which defines models of the KB and gives a meaning to expressions in the KB. Once the problems are formalized, one can start both a theoretical analysis of them, and—maybe independently—a search for reasoning procedures accomplishing the tasks. Completeness and correctness of procedures can be judged with respect to the formal statements of the problems.

Up to now, all the proposed systems give incomplete procedures for solving the above problems 1–4, except for KRIS[1]. That is, some inferences are missed, in some cases without a precise semantical characterization of which ones are. If the designer or the user needs (more) complete reasoning, she/he must either write programs in a suitable programming language (as in the database proposal of Abrial, and in TAXIS), or define appropriate inference rules completing the inference capabilities of the system (as in BACK, LOOM, and CLASSIC). From the theoretical point of view, for several systems (e.g., LOOM) it is not even known if complete procedures can ever exist—i.e., the decidability of the corresponding problems is not known.

Recent research on the computational complexity of subsumption had an influence in many TKRSs on the choice for incomplete procedures. Brachman and Levesque (1984) started this research analyzing the complexity of subsumption between pure concept expressions, abstracting from KBs (we call this problem later in the paper as *pure subsumption*). The motivation for focusing on such a small problem was that pure subsumption is a fundamental inference in any TKRS. It turned out that pure subsumption is tractable (i.e., worst-case polynomial-time solvable) for simple languages, and intractable for slight extensions of such languages, as subsequent research definitely confirmed (Nebel, 1988; Donini, Lenzerini, Nardi, & Nutt, 1991a, 1991b; Schmidt-Schauß & Smolka, 1991; Donini, Hollunder, Lenzerini, Marchetti Spaccamela, Nardi, & Nutt, 1992). Also, beyond computational complexity, pure subsumption was proved undecidable in the TKRSs $\mathcal{U}$ (Schild, 1988), KL-ONE (Schmidt-Schauß, 1989) and NIKL (Patel-Schneider, 1989).

---

1. Also the system CLASSIC is complete, but only w.r.t. a non-standard semantics for the treatment of individuals. Complete reasoning w.r.t. standard semantics for individuals is not provided, and is coNP-hard (Lenzerini & Schaerf, 1991).

Note that extending the language results in enhancing its expressiveness, therefore the result of that research could be summarized as: The more a TKRS language is expressive, the higher is the computational complexity of reasoning in that language—as Levesque (1984) first noted. This result has been interpreted in two different ways, leading to two different TKRSs design philosophies.

1. 'General-purpose languages for TKRSs are intractable, or even undecidable, and tractable languages are not expressive enough to be of practical interest'. Following this interpretation, in several TKRSs (such as NIKL, LOOM and BACK) incomplete procedures for pure subsumption are considered satisfactory (e.g., see (MacGregor & Brill, 1992) for LOOM). Once completeness is abandoned for this basic subproblem, completeness of overall reasoning procedures is not an issue anymore; but other issues arise, such as how to compare incomplete procedures (Heinsohn, Kudenko, Nebel, & Profitlich, 1992), and how to judge a procedure "complete enough" (MacGregor, 1991). As a practical tool, inference rules can be used in such systems to achieve the expected behavior of the KB w.r.t. the information contained in it.

2. 'A TKRS is (by definition) general-purpose, hence it must provide tractable and complete reasoning to a user'. Following this line, other TKRSs (such as KRYPTON and CLASSIC) provide limited tractable languages for expressing concepts, following the "small-can-be-beautiful" approach (see Patel-Schneider, 1984). The gap between what is expressible in the TKRS language and what is needed to be expressed for the application is then filled by the user, by a (sort of) programming with inference rules. Of course, the usual problems present in program development and debugging arise (McGuinness, 1992).

What is common to both approaches is that a user must cope with incomplete reasoning. The difference is that in the former approach, the burden of regaining useful yet missed inferences is mostly left to the developers of the TKRS (and the user is supposed to specify what is "complete enough"), while in the latter this is mainly left to the user. These are perfectly reasonable approaches in a practical context, where incomplete procedures and specialized programs are often used to deal with intractable problems. In our opinion incomplete procedures are just a provisional answer to the problem—the best possible up to now. In order to improve on such an answer, a theoretical analysis of the general problems 1–4 is to be done.

Previous theoretical results do not deal with the problems 1–4 in their full generality. For example, the problems are studied in (Nebel, 1990, Chapter 4), but only incomplete procedures are given, and cycles are not considered. In (Donini, Lenzerini, Nardi, & Schaerf, 1993; Schaerf, 1993a) the complexity of instance checking has been analyzed, but only KBs without a TBox are treated. Instance checking has also been analyzed in (Vilain, 1991), but addressing only that part of the problem which can be performed as parsing.

In addition, we think that the expressiveness of actual systems should be enhanced making terminological cycles (see Nebel, 1990, Chapter 5) available in TKRSs. Such a feature is of undoubtable practical interest (MacGregor, 1992), yet most present TKRSs can only approximate cycles, by using forward inference rules (as in BACK, CLASSIC, LOOM). In our opinion, in order to make terminological cycles fully available in complete TKRSs, a theoretical investigation is still needed.

Previous theoretical work on cycles was done in (Baader, 1990a, 1990b; Baader, Bürkert, Hollunder, Nutt, & Siekmann, 1990; Dionne, Mays, & Oles, 1992, 1993; Nebel, 1990, 1991; Schild, 1991), but considering KBs formed by the TBox alone. Moreover, these approaches do not deal with number restrictions (except for Nebel, 1990, Section 5.3.5) —a basic feature already provided by TKRSs— and the techniques used do not seem easily extensible to reasoning with ABoxes. We compare in detail several of these works with ours in Section 4.

In this paper, we propose a TKRS equipped with a highly expressive language, including constructors often required in practical applications, and prove decidability of problems 1–4. In particular, our system uses the language $\mathcal{ALCNR}$, which supports general complements of concepts, number restrictions and role conjunction. Moreover, the system allows one to express inclusion statements between general concepts and, as a particular case, terminological cycles. We prove decidability by means of a suitable calculus, which is developed extending the well established framework of constraint systems (see Donini et al., 1991a; Schmidt-Schauß & Smolka, 1991), thus exploiting a uniform approach to reasoning in TKRSs. Moreover, our calculus can easily be turned into a decision procedure.

The paper is organized as follows. In Section 2 we introduce the language, and we give it a Tarski-style extensional semantics, which is the most commonly used. Using this semantics, we establish relationships between problems 1–4 which allow us to concentrate on KB-satisfiability only. In Section 3 we provide a calculus for KB-satisfiability, and show correctness and termination of the calculus. Hence, we conclude that KB-satisfiability is decidable in $\mathcal{ALCNR}$, which is the main result of this paper. In Section 4 we compare our approach with previous results on decidable TKRSs, and we establish the equivalence of general (cyclic) inclusion statements and general concept definitions using the descriptive semantics. Finally, we discuss in detail several practical issues related to our results in Section 5.

## 2. Preliminaries

In this section we first present the basic notions regarding concept languages. Then we describe knowledge bases built up using concept languages, and reasoning services that must be provided for extracting information from such knowledge bases.

### 2.1 Concept Languages

In concept languages, concepts represent the classes of objects in the domain of interest, while roles represent binary relations between objects. Complex concepts and roles can be defined by means of suitable constructors applied to concept names and role names. In particular, concepts and roles in $\mathcal{ALCNR}$ can be formed by means of the following syntax (where $P_i$ (for $i = 1, \ldots, k$) denotes a role name, $C$ and $D$ denote arbitrary concepts, and $R$ an arbitrary role):

$$
\begin{aligned}
C, D \quad \longrightarrow \quad & A \mid & & \text{(concept name)} \\
& \top \mid & & \text{(top concept)} \\
& \bot \mid & & \text{(bottom concept)} \\
& (C \sqcap D) \mid & & \text{(conjunction)} \\
& (C \sqcup D) \mid & & \text{(disjunction)} \\
& \neg C \mid & & \text{(complement)} \\
& \forall R.C \mid & & \text{(universal quantification)} \\
& \exists R.C \mid & & \text{(existential quantification)} \\
& (\geq n\, R) \mid (\leq n\, R) & & \text{(number restrictions)} \\
R \quad \longrightarrow \quad & P_1 \sqcap \cdots \sqcap P_k & & \text{(role conjunction)}
\end{aligned}
$$

When no confusion arises we drop the brackets around conjunctions and disjunctions. We interpret concepts as subsets of a domain and roles as binary relations over a domain. More precisely, an *interpretation* $\mathcal{I} = (\Delta^{\mathcal{I}}, \cdot^{\mathcal{I}})$ consists of a nonempty set $\Delta^{\mathcal{I}}$ (the *domain* of $\mathcal{I}$) and a function $\cdot^{\mathcal{I}}$ (the *extension function* of $\mathcal{I}$), which maps every concept to a subset of $\Delta^{\mathcal{I}}$ and every role to a subset of $\Delta^{\mathcal{I}} \times \Delta^{\mathcal{I}}$. The interpretation of concept names and role names is thus restricted by $A^{\mathcal{I}} \subseteq \Delta^{\mathcal{I}}$, and $P^{\mathcal{I}} \subseteq \Delta^{\mathcal{I}} \times \Delta^{\mathcal{I}}$, respectively. Moreover, the interpretation of complex concepts and roles must satisfy the following equations ($\sharp\{\}$ denotes the cardinality of a set):

$$
\begin{aligned}
\top^{\mathcal{I}} &= \Delta^{\mathcal{I}} \\
\bot^{\mathcal{I}} &= \emptyset \\
(C \sqcap D)^{\mathcal{I}} &= C^{\mathcal{I}} \cap D^{\mathcal{I}} \\
(C \sqcup D)^{\mathcal{I}} &= C^{\mathcal{I}} \cup D^{\mathcal{I}} \\
(\neg C)^{\mathcal{I}} &= \Delta^{\mathcal{I}} \setminus C^{\mathcal{I}} \\
(\forall R.C)^{\mathcal{I}} &= \{ d_1 \in \Delta^{\mathcal{I}} \mid \forall d_2 : (d_1, d_2) \in R^{\mathcal{I}} \rightarrow d_2 \in C^{\mathcal{I}} \} \\
(\exists R.C)^{\mathcal{I}} &= \{ d_1 \in \Delta^{\mathcal{I}} \mid \exists d_2 : (d_1, d_2) \in R^{\mathcal{I}} \wedge d_2 \in C^{\mathcal{I}} \} \\
(\geq n\, R)^{\mathcal{I}} &= \{ d_1 \in \Delta^{\mathcal{I}} \mid \quad \sharp\{ d_2 \mid (d_1, d_2) \in R^{\mathcal{I}} \} \geq n \} \\
(\leq n\, R)^{\mathcal{I}} &= \{ d_1 \in \Delta^{\mathcal{I}} \mid \quad \sharp\{ d_2 \mid (d_1, d_2) \in R^{\mathcal{I}} \} \leq n \} \\
(P_1 \sqcap \cdots \sqcap P_k)^{\mathcal{I}} &= P_1^{\mathcal{I}} \cap \cdots \cap P_k^{\mathcal{I}}
\end{aligned}
\tag{1}
$$

## 2.2 Knowledge Bases

A knowledge base built by means of concept languages is generally formed by two components: The *intensional* one, called TBox, and the *extensional* one, called ABox.

We first turn our attention to the TBox. As we said before, the intensional level specifies the properties of the concepts of interest in a particular application. Syntactically, such properties are expressed in terms of what we call *inclusion statements*. An inclusion statement (or simply inclusion) has the form

$$C \sqsubseteq D$$

where $C$ and $D$ are two arbitrary $\mathcal{ALCNR}$-concepts. Intuitively, the statement specifies that every instance of $C$ is also an instance of $D$. More precisely, an interpretation $\mathcal{I}$ *satisfies* the inclusion $C \sqsubseteq D$ if $C^{\mathcal{I}} \subseteq D^{\mathcal{I}}$.

A TBox is a finite set of inclusions. An interpretation $\mathcal{I}$ is a *model* for a TBox $\mathcal{T}$ if $\mathcal{I}$ satisfies all inclusions in $\mathcal{T}$.

In general, TKRSs provide the user with mechanisms for stating *concept introductions* (e.g., Nebel, 1990, Section 3.2) of the form $A \doteq D$ (concept definition, interpreted as set equality), or $A \stackrel{.}{\leq} D$ (concept specification, interpreted as set inclusion), with the restrictions that the left-hand side concept $A$ must be a concept name, that for each concept name at most one introduction is allowed, and that no *terminological cycles* are allowed, i.e., no concept name may occur—neither directly nor indirectly—within its own introduction. These restrictions make it possible to substitute an occurrence of a defined concept by its definition.

We do not impose any of these restrictions to the form of inclusions, obtaining statements that are syntactically more expressive than concept introductions. In particular, a definition of the form $A \doteq D$ can be expressed in our system using the pair of inclusions $A \sqsubseteq D$ and $D \sqsubseteq A$ and a specification of the form $A \stackrel{.}{\leq} D$ can be simply expressed by $A \sqsubseteq D$. Conversely, an inclusion of the form $C \sqsubseteq D$, where $C$ and $D$ are arbitrary concepts, cannot be expressed with concept introductions. Moreover, cyclic inclusions are allowed in our statements, realizing terminological cycles.

As shown in (Nebel, 1991), there are at least three types of semantics for terminological cycles, namely the least fixpoint, the greatest fixpoint, and the descriptive semantics. Fixpoint semantics choose particular models among the set of interpretations that satisfy a statement of the form $A \doteq D$. Such models are chosen as the least and the greatest fixpoint of the above equation. The descriptive semantics instead considers all interpretations that satisfy the statement (i.e., all fixpoints) as its models.

However, fixpoint semantics naturally apply only to fixpoint statements like $A \doteq D$, where $D$ is a "function" of $A$, i.e., $A$ may appear in $D$, and there is no obvious way to extend them to general inclusions. In addition, since our language includes the constructor for complement of general concepts, the "function" $D$ may be not monotone, and therefore the least and the greatest fixpoints may be not unique. Whether there exists or not a definitional semantics that is suitable for cyclic definitions in expressive languages is still unclear.

Conversely, the descriptive semantics interprets statements as just restricting the set of possible models, with no definitional import. Although it is not completely satisfactory in all practical cases (Baader, 1990b; Nebel, 1991), the descriptive semantics has been considered to be the most appropriate one for general cyclic statements in powerful concept languages. Hence, it seems to be the most suitable to be extended to our case and it is exactly the one we have adopted above.

Observe that our decision to put general inclusions in the TBox is not a standard one. In fact, in TKRS like KRYPTON such statements were put in the ABox. However, we conceive inclusions as a generalization of traditional TBox statements: acyclic concept introductions, with their definitional import, can be perfectly expressed with inclusions; and cyclic concept introductions can be expressed as well, if descriptive semantics is adopted. Therefore, we believe that inclusions should be part of the TBox.

Notice that role conjunction allows one to express the practical feature of *subroles*. For example, the role `ADOPTEDCHILD` can be written as `CHILD ⊓ ADOPTEDCHILD′`, where `ADOPTED-CHILD′` is a role name, making it a subrole of `CHILD`. Following such idea, every hierarchy of role names can be rephrased with a set of role conjunctions, and vice versa.

Actual systems usually provide for the construction of hierarchies of roles by means of role introductions (i.e., statements of the form $P \doteq R$ and $P \dot{\le} R$) in the TBox. However, in our simple language for roles, cyclic definitions of roles can be always reduced to acyclic definitions, as explained in (Nebel, 1990, Sec.5.3.1). When role definitions are acyclic, one can always substitute in every concept each role name with its definition, obtaining an equivalent concept. Therefore, we do not consider role definitions in this paper, and we conceive the TBox just as a set of concept inclusions.

Even so, it is worth to notice that concept inclusions can express knowledge about roles. In particular, domain and range restrictions of roles can be expressed, in a way similar to the one in (Catarci & Lenzerini, 1993). Restricting the domain of a role $R$ to a concept $C$ and its range to a concept $D$ can be done by the two inclusions

$$\exists R.\top \sqsubseteq C, \quad \top \sqsubseteq \forall R.D$$

It is straightforward to show that if an interpretation $\mathcal{I}$ satisfies the two inclusions, then $R^{\mathcal{I}} \subseteq C^{\mathcal{I}} \times D^{\mathcal{I}}$.

Combining subroles with domain and range restrictions it is also possible to partially express the constructor for *role restriction*, which is present in various proposals (e.g., the language $\mathcal{FL}$ in Brachman & Levesque, 1984). Role restriction, written as $R : C$, is defined by $(R : C)^{\mathcal{I}} = \{(d_1, d_2) \in \Delta^{\mathcal{I}} \times \Delta^{\mathcal{I}} \mid (d_1, d_2) \in R^{\mathcal{I}} \wedge d_2 \in C^{\mathcal{I}}\}$. For example the role `DAUGHTER`, which can be formulated as `CHILD:Female`, can be partially simulated by `CHILD ⊓ DAUGHTER′`, with the inclusion $\top \sqsubseteq \forall$`DAUGHTER′`.`Female`. However, this simulation would not be complete in number restrictions: E.g., if a mother has at least three daughters, then we know she has at least three female children; if instead we know that she has three female children we cannot infer that she has three daughters.

We can now turn our attention to the *extensional level*, i.e., the ABox. The ABox essentially allows one to specify instance-of relations between individuals and concepts, and between pairs of individuals and roles.

Let $\mathcal{O}$ be an alphabet of symbols, called *individuals*. Instance-of relationships are expressed in terms of *membership assertions* of the form:

$$C(a), \qquad R(a, b),$$

where $a$ and $b$ are individuals, $C$ is an $\mathcal{ALCNR}$-concept, and $R$ is an $\mathcal{ALCNR}$-role. Intuitively, the first form states that $a$ is an instance of $C$, whereas the second form states that $a$ is related to $b$ by means of the role $R$.

In order to assign a meaning to membership assertions, the extension function $\cdot^{\mathcal{I}}$ of an interpretation $\mathcal{I}$ is extended to individuals by mapping them to elements of $\Delta^{\mathcal{I}}$ in such a way that $a^{\mathcal{I}} \neq b^{\mathcal{I}}$ if $a \neq b$. This property is called *Unique Name Assumption*; it ensures that different individuals are interpreted as different objects.

An interpretation $\mathcal{I}$ *satisfies* the assertion $C(a)$ if $a^{\mathcal{I}} \in C^{\mathcal{I}}$, and *satisfies* $R(a, b)$ if $(a^{\mathcal{I}}, b^{\mathcal{I}}) \in R^{\mathcal{I}}$. An ABox is a finite set of membership assertions. $\mathcal{I}$ is a *model* for an ABox $\mathcal{A}$ if $\mathcal{I}$ satisfies all the assertions in $\mathcal{A}$.

An $\mathcal{ALCNR}$-*knowledge base* $\Sigma$ is a pair $\Sigma = \langle \mathcal{T}, \mathcal{A} \rangle$ where $\mathcal{T}$ is a TBox and $\mathcal{A}$ is an ABox. An interpretation $\mathcal{I}$ is a *model* for $\Sigma$ if it is both a model for $\mathcal{T}$ and a model for $\mathcal{A}$.

We can now formally define the problems 1–4 mentioned in the introduction. Let $\Sigma$ be an $\mathcal{ALCNR}$-knowledge base.

1. *KB-satisfiability* : $\Sigma$ is *satisfiable*, if it has a model;

2. *Concept Satisfiability* : $C$ is *satisfiable* w.r.t $\Sigma$, if there exists a model $\mathcal{I}$ of $\Sigma$ such that $C^{\mathcal{I}} \neq \emptyset$;

3. *Subsumption* : $C$ is *subsumed* by $D$ w.r.t. $\Sigma$, if $C^{\mathcal{I}} \subseteq D^{\mathcal{I}}$ for every model $\mathcal{I}$ of $\Sigma$;

4. *Instance Checking* : $a$ is an instance of $C$, written $\Sigma \models C(a)$, if the assertion $C(a)$ is satisfied in every model of $\Sigma$.

In (Nebel, 1990, Sec.3.3.2) it is shown that the ABox plays no active role when checking concept satisfiability and subsumption. In particular, Nebel shows that the ABox (subject to its satisfiability) can be replaced by an empty one without affecting the result of those services. Actually, in (Nebel, 1990), the above property is stated for a language less expressive than $\mathcal{ALCNR}$. However, it is easy to show that it extends to $\mathcal{ALCNR}$. It is important to remark that such a property is not valid for all concept languages. In fact, there are languages that include some constructors that refer to the individuals in the concept language, e.g., the constructor ONE-OF (Borgida et al., 1989) that forms a concept from a set of enumerated individuals. If a concept language includes such a constructor the individuals in the TBox can interact with the individuals in the ABox, as shown in (Schaerf, 1993b). As a consequence, both concept satisfiability and subsumption depend also on the ABox.

**Example 2.1** Consider the following knowledge base $\Sigma = \langle \mathcal{T}, \mathcal{A} \rangle$:

$\mathcal{T} = \{\exists \texttt{TEACHES.Course} \sqsubseteq (\texttt{Student} \sqcap \exists \texttt{DEGREE.BS}) \sqcup \texttt{Prof},$
$\quad\quad \texttt{Prof} \sqsubseteq \exists \texttt{DEGREE.MS},$
$\quad\quad \exists \texttt{DEGREE.MS} \sqsubseteq \exists \texttt{DEGREE.BS},$
$\quad\quad \texttt{MS} \sqcap \texttt{BS} \sqsubseteq \bot\}$
$\mathcal{A} = \{\texttt{TEACHES(john, cs156)}, (\leq 1\, \texttt{DEGREE})(\texttt{john}), \texttt{Course(cs156)}\}$

$\Sigma$ is a fragment of a hypothetical knowledge base describing the organization of a university. The first inclusion, for instance, states that the persons teaching a course are either graduate students (students with a BS degree) or professors. It is easy to see that $\Sigma$ is satisfiable. For example, the following interpretation $\mathcal{I}$ satisfies all the inclusions in $\mathcal{T}$ and all the assertions in $\mathcal{A}$, and therefore it is a model for $\Sigma$:

$\Delta^{\mathcal{I}} = \{\texttt{john}, \texttt{cs156}, \texttt{csb}\}, \ \texttt{john}^{\mathcal{I}} = \texttt{john}, \ \texttt{cs156}^{\mathcal{I}} = \texttt{cs156}$
$\texttt{Student}^{\mathcal{I}} = \{\texttt{john}\}, \ \texttt{Prof}^{\mathcal{I}} = \emptyset, \ \texttt{Course}^{\mathcal{I}} = \{\texttt{cs156}\}, \ \texttt{BS}^{\mathcal{I}} = \{\texttt{csb}\}$
$\texttt{MS}^{\mathcal{I}} = \emptyset, \ \texttt{TEACHES}^{\mathcal{I}} = \{(\texttt{john}, \texttt{cs156})\}, \ \texttt{DEGREE}^{\mathcal{I}} = \{(\texttt{john}, \texttt{csb})\}$

We have described the interpretation $\mathcal{I}$ by giving only $\Delta^{\mathcal{I}}$, and the values of $\mathcal{I}$ on concept names and role names. It is straightforward to see that all values of $\mathcal{I}$ on complex concepts and roles are uniquely determined by imposing that $\mathcal{I}$ must satisfy the Equations 1 on page 113.

Notice that it is possible to draw several non-trivial conclusions from $\Sigma$. For example, we can infer that $\Sigma \models \texttt{Student(john)}$. Intuitively this can be shown as follows: John teaches a course, thus he is either a student with a BS or a professor. But he can't be a professor since professors have at least two degrees (BS and MS) and he has at most one, therefore he is a student.  $\square$

Given the previous semantics, the problems 1–4 can all be reduced to KB-satisfiability (or to its complement) in linear time. In fact, given a knowledge base $\Sigma = \langle \mathcal{T}, \mathcal{A} \rangle$, two concepts $C$ and $D$, an individual $a$, and an individual $b$ not appearing in $\Sigma$, the following equivalences hold:

$C$ is satisfiable w.r.t $\Sigma$   iff   $\langle \mathcal{T}, \mathcal{A} \cup \{C(b)\} \rangle$ is satisfiable.

$C$ is subsumed by $D$ w.r.t. $\Sigma$   iff   $\langle \mathcal{T}, \mathcal{A} \cup \{(C \sqcap \neg D)(b)\} \rangle$ is not satisfiable.

$\Sigma \models C(a)$   iff   $\langle \mathcal{T}, \mathcal{A} \cup \{(\neg C)(a)\} \rangle$ is not satisfiable.

A slightly different form of these equivalences has been given in (Hollunder, 1990). The equivalences given here are a straightforward consequence of the ones given by Hollunder. However, the above equivalences are not valid for languages including constructors that refer to the individuals in the concept language. The equivalences between reasoning services in such languages are studied in (Schaerf, 1993b).

Based on the above equivalences, in the next section we concentrate just on KB-satisfiability.

## 3. Decidability Result

In this section we provide a calculus for deciding KB-satisfiability. In particular, in Subsection 3.1 we present the calculus and we state its correctness. Then, in Subsection 3.2, we prove the termination of the calculus. This will be sufficient to assess the decidability of all problems 1–4, thanks to the relationships between the four problems.

### 3.1 The calculus and its correctness

Our method makes use of the notion of *constraint system* (Donini et al., 1991a; Schmidt-Schauß & Smolka, 1991; Donini, Lenzerini, Nardi, & Schaerf, 1991c), and is based on a tableaux-like calculus (Fitting, 1990) that tries to build a model for the logical formula corresponding to a KB.

We introduce an alphabet of variable symbols $\mathcal{V}$ together with a well-founded total ordering '$\prec$' on $\mathcal{V}$. The alphabet $\mathcal{V}$ is disjoint from the other ones defined so far. The purpose of the ordering will become clear later. The elements of $\mathcal{V}$ are denoted by the letters $x, y, z, w$. From this point on, we use the term *object* as an abstraction for individual and variable (i.e., an object is an element of $\mathcal{O} \cup \mathcal{V}$). Objects are denoted by the symbols $s, t$ and, as in Section 2, individuals are denoted by $a, b$.

A *constraint* is a syntactic entity of one of the forms:

$$s\!:\!C, \quad sPt, \quad \forall x.x\!:\!C, \quad s \not\doteq t,$$

where $C$ is a concept and $P$ is a role name. Concepts are assumed to be *simple*, i.e., the only complements they contain are of the form $\neg A$, where $A$ is a concept name. Arbitrary $\mathcal{ALCNR}$-concepts can be rewritten into equivalent simple concepts in linear time (Donini et al., 1991a). A constraint system is a finite nonempty set of constraints.

Given an interpretation $\mathcal{I}$, we define an $\mathcal{I}$-*assignment* $\alpha$ as a function that maps every variable of $\mathcal{V}$ to an element of $\Delta^{\mathcal{I}}$, and every individual $a$ to $a^{\mathcal{I}}$ (i.e., $\alpha(a) = a^{\mathcal{I}}$ for all $a \in \mathcal{O}$).

A pair $(\mathcal{I}, \alpha)$ *satisfies* the constraint $s\!:\!C$ if $\alpha(s) \in C^{\mathcal{I}}$, the constraint $sPt$ if $(\alpha(s), \alpha(t)) \in P^{\mathcal{I}}$, the constraint $s \not\doteq t$ if $\alpha(s) \neq \alpha(t)$, and finally, the constraint $\forall x.x\!:\!C$ if $C^{\mathcal{I}} = \Delta^{\mathcal{I}}$ (notice that $\alpha$ does not play any role in this case). A constraint system $S$ is *satisfiable* if there is a pair $(\mathcal{I}, \alpha)$ that satisfies every constraint in $S$.

An $\mathcal{ALCNR}$-knowledge base $\Sigma = \langle \mathcal{T}, \mathcal{A} \rangle$ can be translated into a constraint system $S_\Sigma$ by replacing every inclusion $C \sqsubseteq D \in \mathcal{T}$ with the constraint $\forall x.x\!:\!\neg C \sqcup D$, every membership assertion $C(a)$ with the constraint $a\!:\!C$, every $R(a,b)$ with the constraints $aP_1b, \ldots, aP_kb$ if $R = P_1 \sqcap \ldots \sqcap P_k$, and including the constraint $a \not\doteq b$ for every pair $(a,b)$ of individuals appearing in $\mathcal{A}$. It is easy to see that $\Sigma$ is satisfiable if and only if $S_\Sigma$ is satisfiable.

In order to check a constraint system $S$ for satisfiability, our technique adds constraints to $S$ until either an evident contradiction is generated or an interpretation satisfying it can be obtained from the resulting system. Constraints are added on the basis of a suitable set of so-called *propagation rules*.

Before providing the rules, we need some additional definitions. Let $S$ be a constraint system and $R = P_1 \sqcap \ldots \sqcap P_k$ $(k \geq 1)$ be a role. We say that $t$ is an $R$-*successor of $s$ in $S$* if $sP_1t, \ldots, sP_kt$ are in $S$. We say that $t$ is a *direct successor of $s$ in $S$* if for some role $R$, $t$ is an $R$-successor of $s$. We call direct predecessor the inverse relation of direct successor. If $S$ is clear from the context we omit it. Moreover, we denote by *successor* the transitive closure of the relation direct successor, and we denote by *predecessor* its inverse.

We assume that variables are introduced in a constraint system according to the ordering '$\prec$'. This means, if $y$ is introduced in a constraint system $S$ then $x \prec y$ for all variables $x$ that are already in $S$.

We denote by $S[x/s]$ the constraint system obtained from $S$ by replacing each occurrence of the variable $x$ by the object $s$.

We say that $s$ and $t$ are *separated in $S$* if the constraint $s \not\doteq t$ is in $S$.

Given a constraint system $S$ and an object $s$, we define the function $\sigma(\cdot, \cdot)$ as follows: $\sigma(S, s) := \{C \mid s\!:\!C \in S\}$. Moreover, we say that two variables $x$ and $y$ are $S$-*equivalent*, written $x \equiv_s y$, if $\sigma(S, x) = \sigma(S, y)$. Intuitively, two S-equivalent variables can represent the same element in the potential interpretation built by the rules, unless they are separated.

The *propagation rules* are:

1. $S \rightarrow_{\sqcap} \{s{:}C_1, \ s{:}C_2\} \cup S$

   if 1. $s{:}C_1 \sqcap C_2$ is in $S$,
       2. $s{:}C_1$ and $s{:}C_2$ are not both in $S$

2. $S \rightarrow_{\sqcup} \{s{:}D\} \cup S$

   if 1. $s{:}C_1 \sqcup C_2$ is in $S$,
       2. neither $s{:}C_1$ nor $s{:}C_2$ is in $S$,
       3. $D = C_1$ or $D = C_2$

3. $S \rightarrow_{\forall} \{t{:}C\} \cup S$

   if 1. $s{:}\forall R.C$ is in $S$,
       2. $t$ is an $R$-successor of $s$,
       3. $t{:}C$ is not in $S$

4. $S \rightarrow_{\exists} \{sP_1y, \ldots, sP_ky, \ y{:}C\} \cup S$

   if 1. $s{:}\exists R.C$ is in $S$,
       2. $R = P_1 \sqcap \ldots \sqcap P_k$,
       3. $y$ is a new variable,
       4. there is no $t$ such that $t$ is an $R$-successor of $s$ in $S$ and $t{:}C$ is in $S$,
       5. if $s$ is a variable there is no variable $w$ such that $w \prec s$ and $s \equiv_s w$

5. $S \rightarrow_{\geq} \{sP_1y_i, \ldots, sP_ky_i \mid i \in 1..n\} \cup \{y_i \neq y_j \mid i, j \in 1..n, i \neq j\} \cup S$

   if 1. $s{:}(\geq n\ R)$ is in $S$,
       2. $R = P_1 \sqcap \ldots \sqcap P_k$,
       3. $y_1, \ldots, y_n$ are new variables,
       4. there do not exist $n$ pairwise separated $R$-successors of $s$ in $S$,
       5. if $s$ is a variable there is no variable $w$ such that $w \prec s$ and $s \equiv_s w$

6. $S \rightarrow_{\leq} S[y/t]$

   if 1. $s{:}(\leq n\ R)$ is in $S$,
       2. $s$ has more than $n$ $R$-successors in $S$,
       3. $y, t$ are two $R$-successors of $s$ which are not separated

7. $S \rightarrow_{\forall_x} \{s{:}C\} \cup S$

   if 1. $\forall x.x{:}C$ is in $S$,
       2. $s$ appears in $S$,
       3. $s{:}C$ is not in $S$.

We call the rules $\rightarrow_{\sqcup}$ and $\rightarrow_{\leq}$ *nondeterministic* rules, since they can be applied in different ways to the same constraint system (intuitively, they correspond to branching rules of tableaux). All the other rules are called *deterministic* rules. Moreover, we call the rules $\rightarrow_{\exists}$ and $\rightarrow_{\geq}$ *generating* rules, since they introduce new variables in the constraint system. All other rules are called *nongenerating* ones.

The use of the condition based on the $S$-equivalence relation in the generating rules (condition 5) is related to the goal of keeping the constraint system finite even in presence of potentially infinite chains of applications of generating rules. Its role will become clearer later in the paper.

One can verify that rules are always applied to a system $S$ either because of the presence in $S$ of a given constraint $s\!:\!C$ (condition 1), or, in the case of the $\rightarrow_{\forall_x}$-rule, because of the presence of an object $s$ in $S$. When no confusion arises, we will say that a rule is *applied to* the constraint $s\!:\!C$ or the object $s$ (instead of saying that it is applied to the constraint system $S$).

**Proposition 3.1 (Invariance)** *Let $S$ and $S'$ be constraint systems. Then:*

1. *If $S'$ is obtained from $S$ by application of a deterministic rule, then $S$ is satisfiable if and only if $S'$ is satisfiable.*

2. *If $S'$ is obtained from $S$ by application of a nondeterministic rule, then $S$ is satisfiable if $S'$ is satisfiable. Conversely, if $S$ is satisfiable and a nondeterministic rule is applicable to an object $s$ in $S$, then it can be applied to $s$ in such a way that it yields a satisfiable constraint system.*

*Proof.* The proof is mainly a rephrasing of typical soundness proofs for tableaux methods (e.g., Fitting, 1990, Lemma 6.3.2). The only non-standard constructors are number restrictions.

1. "$\Leftarrow$" Considering the deterministic rules one can directly check that $S$ is a subset of $S'$. So it is obvious that $S$ is satisfiable if $S'$ is satisfiable.

"$\Rightarrow$" In order to show that $S'$ is satisfiable if this is the case for $S$ we consider in turn each possible deterministic rule application leading from $S$ to $S'$. We assume that $(\mathcal{I}, \alpha)$ satisfies $S$.

If the $\rightarrow_\sqcap$-rule is applied to $s\!:\!C_1 \sqcap C_2$ in $S$, then $S' = S \cup \{s\!:\!C_1,\ s\!:\!C_2\}$. Since $(\mathcal{I}, \alpha)$ satisfies $s\!:\!C_1 \sqcap C_2$, $(\mathcal{I}, \alpha)$ satisfies $s\!:\!C_1$ and $s\!:\!C_2$ and therefore $S'$.

If the $\rightarrow_\forall$-rule is applied to $s\!:\!\forall R.C$, there must be an $R$-successor $t$ of $s$ in $S$ such that $S' = S \cup \{t\!:\!C\}$. Since $(\mathcal{I}, \alpha)$ satisfies $S$, it holds that $(\alpha(s), \alpha(t)) \in R^\mathcal{I}$. Since $(\mathcal{I}, \alpha)$ satisfies $s\!:\!\forall R.C$, it holds that $\alpha(t) \in C^\mathcal{I}$. So $(\mathcal{I}, \alpha)$ satisfies $t\!:\!C$ and therefore $S'$.

If the $\rightarrow_{\forall_x}$-rule is applied to an $s$ because of the presence of $\forall x.x\!:\!C$ in $S$, then $S' = S \cup \{s\!:\!C\}$. Since $(\mathcal{I}, \alpha)$ satisfies $S$ it holds that $C^\mathcal{I} = \Delta^\mathcal{I}$. Therefore $\alpha(s) \in C^\mathcal{I}$ and so $(\mathcal{I}, \alpha)$ satisfies $S'$.

If the $\rightarrow_\exists$-rule is applied to $s\!:\!\exists R.C$, then $S' = S \cup \{sP_1y, \dots, sP_ky,\ y\!:\!C\}$. Since $(\mathcal{I}, \alpha)$ satisfies $S$, there exists a $d$ such that $(\alpha(s), d) \in R^\mathcal{I}$ and $d \in C^\mathcal{I}$. We define the $\mathcal{I}$-assignment $\alpha'$ as $\alpha'(y) := d$ and $\alpha'(t) := \alpha(t)$ for $t \neq y$. It is easy to show that $(\mathcal{I}, \alpha')$ satisfies $S'$.

If the $\rightarrow_\geq$-rule is applied to $s\!:\!(\geq n\,R)$, then $S' = S \cup \{sP_1y_i, \dots, sP_ky_i \mid i \in 1..n\} \cup \{y_i \not\doteq y_j \mid i, j \in 1..n, i \neq j\}$. Since $(\mathcal{I}, \alpha)$ satisfies $S$, there exist $n$ distinct elements $d_1, \dots, d_n \in \Delta^\mathcal{I}$ such that $(\alpha(s), d_i) \in R^\mathcal{I}$. We define the $\mathcal{I}$-assignment $\alpha'$ as $\alpha'(y_i) := d_i$ for $i \in 1..n$ and $\alpha'(t) := \alpha(t)$ for $t \notin \{y_1, \dots, y_n\}$. It is easy to show that $(\mathcal{I}, \alpha')$ satisfies $S'$.

2. "$\Leftarrow$" Assume that $S'$ is satisfied by $(\mathcal{I}, \alpha')$. We show that $S$ is also satisfiable. If $S'$ is obtained from $S$ by application of the $\rightarrow_\sqcup$-rule, then $S$ is a subset of $S'$ and therefore satisfied by $(\mathcal{I}, \alpha')$.

If $S'$ is obtained from $S$ by application of the $\rightarrow_\leq$-rule to $s{:}(\leq n\, R)$ in $S$, then there are $y, t$ in $S$ such that $S' = S[y/t]$. We define the $\mathcal{I}$-assignment $\alpha$ as $\alpha(y) := \alpha'(t)$ and $\alpha(v) := \alpha'(v)$ for every object $v$ with $v \neq y$. Obviously $(\mathcal{I}, \alpha)$ satisfies $S$.

"$\Rightarrow$" Now suppose that $S$ is satisfied by $(\mathcal{I}, \alpha)$ and a nondeterministic rule is applicable to an object $s$.

If the $\rightarrow_\sqcup$-rule is applicable to $s{:}C_1 \sqcup C_2$ then, since $S$ is satisfiable, $\alpha(s) \in (C_1 \sqcup C_2)^\mathcal{I}$. It follows that either $\alpha(s) \in C_1^\mathcal{I}$ or $\alpha(s) \in C_2^\mathcal{I}$ (or both). Hence, the $\rightarrow_\sqcup$-rule can obviously be applied in a way such that $(\mathcal{I}, \alpha)$ satisfies the resulting constraint system $S'$.

If the $\rightarrow_\leq$-rule is applicable to $s{:}(\leq n\, R)$, then—since $(\mathcal{I}, \alpha)$ satisfies $S$—it holds that $\alpha(s) \in (\leq n\, R)^\mathcal{I}$ and therefore the set $\{d \in \Delta^\mathcal{I} \mid (\alpha(s), d) \in R^\mathcal{I}\}$ has at most $n$ elements. On the other hand, there are more than $n$ $R$-successors of $s$ in $S$ and for each $R$-successor $t$ of $s$ we have $(\alpha(s), \alpha(t)) \in R^\mathcal{I}$. Thus, we can conclude by the Pigeonhole Principle (see e.g., Lewis & Papadimitriou, 1981, page 26) that there exist at least two $R$-successors $t, t'$ of $s$ such that $\alpha(t) = \alpha(t')$. Since $(\mathcal{I}, \alpha)$ satisfies $S$, the constraint $t \neq t'$ is not in $S$. Therefore one of the two must be a variable, let's say $t' = y$. Now obviously $(\mathcal{I}, \alpha)$ satisfies $S[y/t]$. $\square$

Given a constraint system $S$, more than one rule might be applicable to it. We define the following *strategy* for the application of rules:

1. apply a rule to a variable only if no rule is applicable to individuals;

2. apply a rule to a variable $x$ only if no rule is applicable to a variable $y$ such that $y \prec x$;

3. apply generating rules only if no nongenerating rule is applicable.

The above strategy ensures that the variables are processed one at a time according to the ordering '$\prec$'.

From this point on, we assume that rules are always applied according to this strategy and that we always start with a constraint system $S_\Sigma$ coming from an $\mathcal{ALCNR}$-knowledge base $\Sigma$. The following lemma is a direct consequence of these assumptions.

**Lemma 3.2 (Stability)** *Let $S$ be a constraint system and $x$ be a variable in $S$. Let a generating rule be applicable to $x$ according to the strategy. Let $S'$ be any constraint system derivable from $S$ by any sequence (possibly empty) of applications of rules. Then*

1. *No rule is applicable in $S'$ to a variable $y$ with $y \prec x$*

2. $\sigma(S, x) = \sigma(S', x)$

3. *If $y$ is a variable in $S$ with $y \prec x$ then $y$ is a variable in $S'$, i.e., the variable $y$ is not substituted by another variable or by a constant.*

*Proof.* 1. By contradiction: Suppose $S \equiv S_0 \rightarrow_* S_1 \rightarrow_* \cdots \rightarrow_* S_n \equiv S'$, where $* \in \{\sqcup, \sqcap, \exists, \forall, \geq, \leq, \forall x\}$ and a rule is applicable to a variable $y$ such that $y \prec x$ in $S'$. Then there exists a minimal $i$, where $i \leq n$, such that this is the case in $S_i$. Note that $i \neq 0$; in fact, because of the strategy, if a rule is applicable to $x$ in $S$ no rule is applicable to $y$ in $S$. So no rule is applicable to any variable $z$ such that $z \prec x$ in $S_0, \ldots, S_{i-1}$. It follows that from $S_{i-1}$ to $S_i$ a rule is applied to $x$ or to a variable $w$ such that $x \prec w$. By an exhaustive analysis of all rules we see that—whichever is the rule applied from $S_{i-1}$ to $S_i$—no new constraint of the form $y\!:\!C$ or $yRz$ can be added to $S_{i-1}$, and therefore no rule is applicable to $y$ in $S_i$, contradicting the assumption.

2. By contradiction: Suppose $\sigma(S, x) \neq \sigma(S', x)$. Call $y$ the direct predecessor of $x$, then a rule must have been applied either to $y$ or to $x$ itself. Obviously we have $y \prec x$, therefore the former case cannot be because of point 1. A case analysis shows that the only rules which can have been applied to $x$ are generating ones and the $\rightarrow_\forall$ and the $\rightarrow_\leq$ rules. But these rules add new constraints only to the direct successors of $x$ and not to $x$ itself and therefore do not change $\sigma(\cdot, x)$.

3. This follows from point 1. and the strategy. $\qquad\qquad\square$

Lemma 3.2 proves that for a variable $x$ which has a direct successor, $\sigma(\cdot, x)$ is stable, i.e., it will not change because of subsequent applications of rules. In fact, if a variable has a direct successor it means that a generating rule has been applied to it, therefore (Lemma 3.2.2) from that point on $\sigma(\cdot, x)$ does not change.

A constraint system is *complete* if no propagation rule applies to it. A complete system derived from a system $S$ is also called a *completion* of $S$. A *clash* is a constraint system having one of the following forms:

- $\{s\!:\!\bot\}$

- $\{s\!:\!A,\ s\!:\!\neg A\}$, where $A$ is a concept name.

- $\{s\!:\!(\leq n\, R)\} \cup \{sP_1 t_i, \ldots, sP_k t_i \mid i \in 1..n+1\}$
  $\qquad\qquad \cup\, \{t_i \not\approx t_j \mid i, j \in 1..n+1, i \neq j\}$,

  where $R = P_1 \sqcap \ldots \sqcap P_k$.

A clash is evidently an unsatisfiable constraint system. For example, the last case represents the situation in which an object has an at-most restriction and a set of $R$-successors that cannot be identified (either because they are individuals or because they have been created by some at-least restrictions).

Any constraint system containing a clash is obviously unsatisfiable. The purpose of the calculus is to generate completions, and look for the presence of clashes inside. If $S$ is a completion of $S_\Sigma$ and $S$ contains no clash, we prove that it is always possible to construct a model for $\Sigma$ on the basis of $S$. Before looking at the technical details of the proof, let us consider an example of application of the calculus for checking satisfiability.

**Example 3.3** Consider the following knowledge base $\Sigma = \langle \mathcal{T}, \mathcal{A} \rangle$:

$\mathcal{T} = \{\texttt{Italian} \sqsubseteq \exists\texttt{FRIEND.Italian}\}$

$\mathcal{A} = \{\texttt{FRIEND(peter, susan)},$
$\qquad \forall\texttt{FRIEND.}\neg\texttt{Italian(peter)},$
$\qquad \exists\texttt{FRIEND.Italian(susan)}\}$

The corresponding constraint system $S_\Sigma$ is:

$S_\Sigma = \{\forall x . x\!:\!\neg\texttt{Italian} \sqcup \exists\texttt{FRIEND.Italian},$
$\qquad \texttt{peterFRIENDsusan},$

```
peter:∀FRIEND.¬Italian,
susan:∃FRIEND.Italian
peter ≠ susan}
```

A sequence of applications of the propagation rules to $S_\Sigma$ is as follows:

$S_1 = S_\Sigma \cup \{\text{susan}: \neg\text{Italian}\}$ ($\rightarrow_\forall$-rule)

$S_2 = S_1 \cup \{\text{peter}: \neg\text{Italian} \sqcup \exists\text{FRIEND.Italian}\}$ ($\rightarrow_{\forall_x}$-rule)

$S_3 = S_2 \cup \{\text{susan}: \neg\text{Italian} \sqcup \exists\text{FRIEND.Italian}\}$ ($\rightarrow_{\forall_x}$-rule)

$S_4 = S_3 \cup \{\text{peter}: \neg\text{Italian}\}$ ($\rightarrow_\sqcup$-rule)

$S_5 = S_4 \cup \{\text{susanFRIENDx}, \text{x}: \text{Italian}\}$ ($\rightarrow_\exists$-rule)

$S_6 = S_5 \cup \{\text{x}: \neg\text{Italian} \sqcup \exists\text{FRIEND.Italian}\}$ ($\rightarrow_{\forall_x}$-rule)

$S_7 = S_6 \cup \{\text{x}: \exists\text{FRIEND.Italian}\}$ ($\rightarrow_\sqcup$-rule)

$S_8 = S_7 \cup \{\text{xFRIENDy}, \text{y}: \text{Italian}\}$ ($\rightarrow_\exists$-rule)

$S_9 = S_8 \cup \{\text{y}: \neg\text{Italian} \sqcup \exists\text{FRIEND.Italian}\}$ ($\rightarrow_{\forall_x}$-rule)

$S_{10} = S_9 \cup \{\text{y}: \exists\text{FRIEND.Italian}\}$ ($\rightarrow_\sqcup$-rule)

One can verify that $S_{10}$ is a complete clash-free constraint system. In particular, the $\rightarrow_\exists$-rule is not applicable to $y$. In fact, since $x \equiv_{S_{10}} y$ condition 5 is not satisfied. From $S_{10}$ one can build an interpretation $\mathcal{I}$, as follows (again, we give only the interpretation of concept and role names):

$\Delta^\mathcal{I} = \{\text{peter}, \text{susan}, \text{x}, \text{y}\}$

$\text{peter}^\mathcal{I} = \text{peter}, \text{susan}^\mathcal{I} = \text{susan}, \alpha(\text{x}) = \text{x}, \alpha(\text{y}) = \text{y},$

$\text{Italian}^\mathcal{I} = \{\text{x}, \text{y}\}$

$\text{FRIEND}^\mathcal{I} = \{(\text{peter}, \text{susan}), (\text{susan}, \text{x}), (\text{x}, \text{y}), (\text{y}, \text{y})\}$

It is easy to see that $\mathcal{I}$ is indeed a model for $\Sigma$. $\square$

In order to prove that it is always possible to obtain an interpretation from a complete clash-free constraint system we need some additional notions. Let $S$ be a constraint system and $x, w$ variables in $S$. We call $w$ a *witness of $x$ in $S$* if the three following conditions hold:

1. $x \equiv_s w$

2. $w \prec x$

3. there is no variable $z$ such that $z \prec w$ and $z$ satisfies conditions 1. and 2., i.e., $w$ is the least variable w.r.t. $\prec$ satisfying conditions 1. and 2.

We say $x$ *is blocked (by $w$)* in $S$ if $x$ has a witness $(w)$ in $S$. The following lemma states a property of witnesses.

**Lemma 3.4** *Let $S$ be a constraint system, $x$ a variable in $S$. If $x$ is blocked then*

1. *$x$ has no direct successor and*

2. *$x$ has exactly one witness.*

*Proof. 1.* By contradiction: Suppose that $x$ is blocked in $S$ and $x P y$ is in $S$. During the completion process leading to $S$ a generating rule must have been applied to $x$ in a system $S'$. It follows from the definition of the rules that in $S'$ for every variable $w \prec x$ we had $x \not\equiv_{S'} w$. Now from Lemma 3.2 we know, that for the constraint system $S$ derivable from $S'$ and for every $w \prec x$ in $S$ we also have $x \not\equiv_{S} w$. Hence there is no witness for $x$ in $S$, contradicting the hypothesis that $x$ is blocked.

*2.* This follows directly from condition 3. for a witness. $\qquad\square$

As a consequence of Lemma 3.4, in a constraint system $S$, if $w_1$ is a witness of $x$ then $w_1$ cannot have a witness itself, since both the relations '$\prec$' and $S$-equivalence are transitive. The uniqueness of the witness for a blocked variable is important for defining the following particular interpretation out of $S$.

Let $S$ be a constraint system. We define the *canonical interpretation* $\mathcal{I}_S$ and the *canonical $\mathcal{I}_S$-assignment* $\alpha_S$ as follows:

1. $\Delta^{\mathcal{I}_S} := \{ s \mid s \text{ is an object in } S \}$

2. $\alpha_S(s) := s$

3. $s \in A^{\mathcal{I}_S}$ if and only if $s{:}A$ is in $S$

4. $(s, t) \in P^{\mathcal{I}_S}$ if and only if

   (a) $s P t$ is in $S$    or

   (b) $s$ is a blocked variable, $w$ is the witness of $s$ in $S$ and $w P t$ is in $S$.

We call $(s, t)$ a *$P$-role-pair* of $s$ in $\mathcal{I}_S$ if $(s, t) \in P^{\mathcal{I}_S}$, we call $(s, t)$ a *role-pair* of $s$ in $\mathcal{I}_S$ if $(s, t)$ is a *$P$-role-pair* for some role $P$. We call a role-pair *explicit* if it comes up from case 4.(a) of the definition of the canonical interpretation and we call it *implicit* if it comes up from case 4.(b).

From Lemma 3.4 it is obvious that a role-pair cannot be both explicit and implicit. Moreover, if a variable has an implicit role-pair then all its role-pairs are implicit and they all come from exactly one witness, as stated by the following lemma.

**Lemma 3.5** *Let $S$ be a completion and $x$ a variable in $S$. Let $\mathcal{I}_S$ be the canonical interpretation for $S$. If $x$ has an implicit role-pair $(x, y)$, then*

1. *all role-pairs of $x$ in $\mathcal{I}_S$ are implicit*

2. *there is exactly one witness $w$ of $x$ in $S$ such that for all roles $P$ in $S$ and all $P$-role-pairs $(x, y)$ of $x$, the constraint $w P y$ is in $S$.*

*Proof.* The first statement follows from Lemma 3.4 (point *1*). The second statement follows from Lemma 3.4 (point *2*) together with the definition of $\mathcal{I}_S$. $\qquad\square$

We have now all the machinery needed to prove the main theorem of this subsection.

**Theorem 3.6** *Let $S$ be a complete constraint system. If $S$ contains no clash then it is satisfiable.*

*Proof.* Let $\mathcal{I}_S$ and $\alpha_S$ be the canonical interpretation and canonical $\mathcal{I}$-assignment for $S$. We prove that the pair $(\mathcal{I}_S, \alpha_S)$ satisfies every constraint $c$ in $S$. If $c$ has the form $sPt$ or $s \neq t$, then $(\mathcal{I}_S, \alpha_S)$ satisfies them by definition of $\mathcal{I}_S$ and $\alpha_S$. Considering the $\rightarrow_\geq$-rule and the $\rightarrow_\leq$-rule we see that a constraint of the form $s \neq s$ can not be in $S$. If $c$ has the form $s: C$, we show by induction on the structure of $C$ that $s \in C^{\mathcal{I}_S}$.

We first consider the base cases. If $C$ is a concept name, then $s \in C^{\mathcal{I}_S}$ by definition of $\mathcal{I}_S$. If $C = \top$, then obviously $s \in \top^{\mathcal{I}_S}$. The case that $C = \bot$ cannot occur since $S$ is clash-free.

Next we analyze in turn each possible complex concept $C$. If $C$ is of the form $\neg C_1$ then $C_1$ is a concept name since all concepts are simple. Then the constraint $s: C_1$ is not in $S$ since $S$ is clash-free. Then $s \notin C_1^{\mathcal{I}_S}$, that is, $s \in \Delta^{\mathcal{I}_S} \setminus C_1^{\mathcal{I}_S}$. Hence $s \in (\neg C_1)^{\mathcal{I}_S}$.

If $C$ is of the form $C_1 \sqcap C_2$ then (since $S$ is complete) $s: C_1$ is in $S$ and $s: C_2$ is in $S$. By induction hypothesis, $s \in C_1^{\mathcal{I}_S}$ and $s \in C_2^{\mathcal{I}_S}$. Hence $s \in (C_1 \sqcap C_2)^{\mathcal{I}_S}$.

If $C$ is of the form $C_1 \sqcup C_2$ then (since $S$ is complete) either $s: C_1$ is in $S$ or $s: C_2$ is in $S$. By induction hypothesis, either $s \in C_1^{\mathcal{I}_S}$ or $s \in C_2^{\mathcal{I}_S}$. Hence $s \in (C_1 \sqcup C_2)^{\mathcal{I}_S}$.

If $C$ is of the form $\forall R.D$, we have to show that for all $t$ with $(s, t) \in R^{\mathcal{I}_S}$ it holds that $t \in D^{\mathcal{I}_S}$. If $(s, t) \in R^{\mathcal{I}_S}$, then according to Lemma 3.5 two cases can occur. Either $t$ is an $R$-successor of $s$ in $S$ or $s$ is blocked by a witness $w$ in $S$ and $t$ is an $R$-successor of $w$ in $S$. In the first case $t: D$ must also be in $S$ since $S$ is complete. Then by induction hypothesis we have $t \in D^{\mathcal{I}_S}$. In the second case by definition of witness, $w: \forall R.D$ is in $S$ and then because of completeness of $S$, $t: D$ must be in $S$. By induction hypothesis we have again $t \in D^{\mathcal{I}_S}$.

If $C$ is of the form $\exists R.D$ we have to show that there exists a $t \in \Delta^{\mathcal{I}_S}$ with $(s, t) \in R^{\mathcal{I}_S}$ and $t \in D^{\mathcal{I}_S}$. Since $S$ is complete, either there is a $t$ that is an $R$-successor of $s$ in $S$ and $t: D$ is in $S$, or $s$ is a variable blocked by a witness $w$ in $S$. In the first case, by induction hypothesis and the definition of $\mathcal{I}_S$, we have $t \in D^{\mathcal{I}_S}$ and $(s, t) \in R^{\mathcal{I}_S}$. In the second case $w: \exists R.D$ is in $S$. Since $w$ cannot be blocked and $S$ is complete, we have that there is a $t$ that is an $R$-successor of $w$ in $S$ and $t: D$ is in $S$. So by induction hypothesis we have $t \in D^{\mathcal{I}_S}$ and by the definition of $\mathcal{I}_S$ we have $(s, t) \in R^{\mathcal{I}_S}$.

If $C$ is of the form $(\leq n R)$ we show the goal by contradiction. Assume that $s \notin (\leq n R)^{\mathcal{I}_S}$. Then there exist atleast $n + 1$ distinct objects $t_1, \ldots, t_{n+1}$ with $(s, t_i) \in R^{\mathcal{I}_S}$, $i \in 1..n + 1$. This means that, since $R = P_1 \sqcap \ldots \sqcap P_k$, there are pairs $(s, t_i) \in P_j^{\mathcal{I}_S}$, where $i \in 1..n + 1$ and $j \in 1..k$. Then according to Lemma 3.5 one of the two following cases must occur. Either all $sP_j t_i$ for $j \in 1..k$, $i \in 1..n + 1$ are in $S$ or there exists a witness $w$ of $s$ in $S$ with all $wP_i t_i$ for $j \in 1..k$ and $i \in 1..n + 1$ are in $S$. In the first case the $\rightarrow_\leq$-rule can not be applicable because of completeness. This means that all the $t_i$'s are pairwise separated, i.e., that $S$ contains the constraints $t_i \neq t_j$, $i, j \in 1..n + 1, i \neq j$. This contradicts the fact that $S$ is clash-free. And the second case leads to an analogous contradiction.

If $C$ is of the form $(\geq n R)$ we show the goal by contradiction. Assume that $s \notin (\geq n R)^{\mathcal{I}_S}$. Then there exist atmost $m < n$ ($m$ possibly 0) distinct objects $t_1, \ldots, t_m$ with $(s, t_i) \in R^{\mathcal{I}_S}$, $i \in 1..m$. We have to consider two cases. First case: $s$ is not blocked in $S$. Since there are only $m$ $R$-successors of $s$ in $S$, the $\rightarrow_\geq$-rule is applicable to $s$. This contradicts the fact that $S$ is complete. Second case: $s$ is blocked by a witness $w$ in $S$. Since there are $m$ $R$-successors of $w$ in $S$, the $\rightarrow_\geq$-rule is applicable to $w$. But this leads to the same contradiction.

If $c$ has the form $\forall x.x\!:\!D$ then, since $S$ is complete, for each object $t$ in $S$, $t\!:\!D$ is in $S$—and, by the previous cases, $t \in D^{\mathcal{I}_S}$. Therefore, the pair $(\mathcal{I}_S, \alpha_S)$ satisfies $\forall x.x\!:\!D$. Finally, since $(\mathcal{I}_S, \alpha_S)$ satisfies all constraints in $S$, $(\mathcal{I}_S, \alpha_S)$ satisfies $S$. $\qquad\square$

**Theorem 3.7 (Correctness)** *A constraint system $S$ is satisfiable if and only if there exists at least one clash-free completion of $S$.*

*Proof.* "$\Leftarrow$" Follows immediately from Theorem 3.6. "$\Rightarrow$" Clearly, a system containing a clash is unsatisfiable. If every completion of $S$ is unsatisfiable, then from Proposition 3.1 $S$, is unsatisfiable. $\qquad\square$

### 3.2 Termination and complexity of the calculus

Given a constraint system $S$, we call $n_S$ the number of concepts appearing in $S$, including also all the concepts appearing as a substring of another concept. Notice that $n_S$ is bounded by the length of the string expressing $S$.

**Lemma 3.8** *Let $S$ be a constraint system and let $S'$ be derived from $S$ by means of the propagation rules. In any set of variables in $S'$ including more than $2^{n_S}$ variables there are at least two variables $x$,$y$ such that $x \equiv_{s'} y$.*

*Proof.* Each constraint $x\!:\!C \in S'$ may contain only concepts of the constraint system $S$. Since there are $n_S$ such concepts, given a variable $x$ there cannot be more than $2^{n_S}$ different sets of constraints $x\!:\!C$ in $S'$. $\qquad\square$

**Lemma 3.9** *Let $S$ be a constraint system and let $S'$ be any constraint system derived from $S$ by applying the propagation rules with the given strategy. Then, in $S'$ there are at most $2^{n_S}$ non-blocked variables.*

*Proof.* Suppose there are $2^{n_S} + 1$ non-blocked variables. From Lemma 3.8, we know that in $S'$ there are at least two variables $y_1$, $y_2$ such that $y_1 \equiv_s y_2$. Obviously either $y_1 \prec y_2$ or $y_2 \prec y_1$ holds; suppose that $y_1 \prec y_2$. From the definitions of witness and blocked either $y_1$ is a witness of $y_2$ or there exists a variable $y_3$ such that $y_3 \prec y_1$ and $y_3$ is a witness of $y_2$. In both cases $y_2$ is blocked, contradicting the hypothesis. $\qquad\square$

**Theorem 3.10 (Termination and space complexity)** *Let $\Sigma$ be an $\mathcal{ALCNR}$-knowledge base and let $n$ be its size. Every completion of $S_\Sigma$ is finite and its size is $O(2^{4n})$.*

*Proof.* Let $S$ be a completion of $S_\Sigma$. From Lemma 3.9 it follows that there are at most $2^n$ non-blocked variables in $S$. Therefore there are at most $m \times 2^n$ total variables in $S$, where $m$ is the maximum number of direct successors for a variable in $S$.

Observe that $m$ is bounded by the number of $\exists R.C$ concepts (at most $n$) plus the sum of all numbers appearing in number restrictions. Since these numbers are expressed in binary, their sum is bounded by $2^n$. Hence, $m \leq 2^n + n$. Since the number of individuals is also bounded by $n$, the total number of objects in $S$ is at most $m \times (2^n + n) \leq (2^n + n) \times (2^n + n)$, that is, $O(2^{2n})$.

The number of different constraints of the form $s:C$, $\forall x.x:C$ in which each object $s$ can be involved is bounded by $n$, and each constraint has size linear in $n$. Hence, the total size of these constraints is bounded by $n \times n \times 2^{2n}$, that is $O(2^{3n})$.

The number of constraints of the form $sPt$, $s \neq t$ is bounded by $(2^{2n})^2 = 2^{4n}$, and each constraint has constant size.

In conclusion, we have that the size of $S$ is $O(2^{4n})$.  □

Notice that the above one is just a coarse upper bound, obtained for theoretical purposes. In practical cases we expect the actual size to be much smaller than that. For example, if the numbers involved in number restrictions were either expressed in unary notation, or limited by a constant (the latter being a reasonable restriction in practical systems) then an argumentation analogous to the above one would lead to a bound of $2^{3n}$.

**Theorem 3.11 (Decidability)** *Given an $\mathcal{ALCNR}$-knowledge base $\Sigma$, checking whether $\Sigma$ is satisfiable is a decidable problem.*

*Proof.* This follows from Theorems 3.7 and 3.10 and the fact that $\Sigma$ is satisfiable if and only if $S_\Sigma$ is satisfiable.  □

We can refine the above theorem, by giving tighter bounds on the time required to decide satisfiability.

**Theorem 3.12 (Time complexity)** *Given an $\mathcal{ALCNR}$-knowledge base $\Sigma$, checking whether $\Sigma$ is satisfiable can be done in nondeterministic exponential time.*

*Proof.* In order to prove the claim it is sufficient to show that each completion is obtained with an exponential number of applications of rules. Since the number of constraints of each completion is exponential (Theorem 3.10) and each rule, but the $\rightarrow_{\leq}$-rule, adds new constraints to the constraint system, it follows that all such rules are applied at most an exponential number of times. Regarding the $\rightarrow_{\leq}$-rule, it is applied for each object at most as many times as the number of its direct successors. Since such number is at most exponential (if numbers are coded in binary) w.r.t. the size of the knowledge base, the claim follows.  □

A lower bound of the complexity of KB-satisfiability is obtained exploiting previous results about the language $\mathcal{ALC}$, which is a sublanguage of $\mathcal{ALCNR}$ that does not include number restrictions and role conjunction. We know from McAllester (1991), and (independently) from an observation by Nutt (1992) that KB-satisfiability in $\mathcal{ALC}$-knowledge bases is EXPTIME-hard (see (Garey & Johnson, 1979, page 183) for a definition) and hence it is hard for $\mathcal{ALCNR}$-knowledge bases, too. Hence, we do not expect to find any algorithm solving the problem in polynomial space, unless PSPACE=EXPTIME. Therefore, we do not expect to substantially improve space complexity of our calculus, which already works in exponential space. We now discuss possible improvements on time complexity.

The proposed calculus works in nondeterministic exponential time, and hence improves the one we proposed in (Buchheit, Donini, & Schaerf, 1993, Sec.4), which works in deterministic double exponential time. The key improvement is that we showed that a KB has a model if and only if it has a model of exponential size. However, it may be argued that as it is, the calculus cannot yet be turned into a practical procedure, since such a procedure would simply simulate nondeterminism by a second level of exponentiality, resulting in a double exponential time procedure. However, the different combinations of concepts are only exponentially many (this is just the cardinality of the powerset of the set of concepts). Hence, a double exponential time procedure wastes most of the time re-analyzing over and over objects with different names yet with the same $\sigma(\cdot, \cdot)$, in different constraint systems. This could be avoided if we allow a variable to be blocked by a witness that is in a *previously analyzed* constraint system. This technique would be similar to the one used in (Pratt, 1978), and to the tree-automata technique used in (Vardi & Wolper, 1986), improving on simple tableaux methods for variants of propositional dynamic logics. Since our calculus considers only one constraint system at a time, a modification of the calculus would be necessary to accomplish this task in a formal way, which is outside the scope of this paper. The formal development of such a deterministic exponential time procedure will be a subject for future research.

Notice that, since the domain of the canonical interpretation $\Delta^{\mathcal{I}_S}$ is always finite, we have also implicitly proved that $\mathcal{ALCNR}$-knowledge bases have the *finite model property*, i.e., any satisfiable knowledge base has a finite model. This property has been extensively studied in modal logics (Hughes & Cresswell, 1984) and dynamic logics (Harel, 1984). In particular, a technique, called *filtration*, has been developed both to prove the finite model property and to build a finite model for a satisfiable formula. This technique allows one to build a finite model from an infinite one by grouping the worlds of a structure in equivalence classes, based on the set of formulae that are satisfied in each world. It is interesting to observe that our calculus, based on witnesses, can be considered as a variant of the filtration technique where the equivalence classes are determined on the basis of our $S$-equivalence relation. However, because of number restrictions, variables that are $S$-equivalent cannot be grouped, since they might be separated (e.g., they might have been introduced by the same application of the $\longrightarrow_{\geq}$-rule). Nevertheless, they can have the same direct successors, as stated in point 4.(b) of the definition of canonical interpretation on page 124. This would correspond to grouping variables of an infinite model in such a way that separations are preserved.

## 4. Relation to previous work

In this section we discuss the relation of our paper to previous work about reasoning with inclusions. In particular, we first consider previously proposed reasoning techniques that deal with inclusions and terminological cycles, then we discuss the relation between inclusions and terminological cycles.

### 4.1 Reasoning Techniques

As mentioned in the introduction, previous results were obtained by Baader et al. (1990), Baader (1990a, 1990b), Nebel (1990, 1991), Schild (1991) and Dionne et al. (1992, 1993).

Nebel (1990, Chapter 5) considers the language $\mathcal{TF}$, containing concept conjunction, universal quantification and number restrictions, and TBoxes containing (possibly cyclic) concept definitions, role definitions and disjointness axioms (stating that two concept names are disjoint). Nebel shows that subsumption of $\mathcal{TF}$-concepts w.r.t. a TBox is decidable. However, the argument he uses is non-constructive: He shows that it is sufficient to con-sider finite interpretations of a size bounded by the size of the TBox in order to decide subsumption.

In (Baader, 1990b) the effect of the three types of semantics—descriptive, greatest fixpoint and least fixpoint semantics—for the language $\mathcal{FL}_0$, containing concept conjunction and universal quantification, is described with the help of *finite automata*. Baader reduces subsumption of $\mathcal{FL}_0$-concepts w.r.t. a TBox containing (possibly cyclic) definitions of the form $A \doteq C$ (which he calls terminological axioms) to decision problems for finite automata. In particular, he shows that subsumption w.r.t. descriptive semantics can be decided in polynomial space using *Büchi automata*. Using results from (Baader, 1990b), in (Nebel, 1991) a characterization of the above subsumption problem w.r.t. descriptive semantics is given with the help of deterministic automata (whereas Büchi automata are nondeterministic). This also yields a PSPACE-algorithm for deciding subsumption.

In (Baader et al., 1990) the attention is restricted to the language $\mathcal{ALC}$. In particular, that paper considers the problem of checking the satisfiability of a single equation of the form $C = \top$, where $C$ is an $\mathcal{ALC}$-concept. This problem, called the *universal satisfiability problem*, is shown to be equivalent to checking the satisfiability of an $\mathcal{ALC}$-TBox (see Proposition 4.1).

In (Baader, 1990a), an extension of $\mathcal{ALC}$, called $\mathcal{ALC}_{reg}$, is introduced, which supports a constructor to express the transitive closure of roles. By means of transitive closure of roles it is possible to replace cyclic inclusions of the form $A \sqsubseteq D$ with equivalent acyclic ones. The problem of checking the satisfiability of an $\mathcal{ALC}_{reg}$-concept is solved in that paper. It is also shown that using transitive closure it is possible to reduce satisfiability of an $\mathcal{ALC}$-concept w.r.t. an $\mathcal{ALC}$-TBox $\mathcal{T} = \{C_1 \sqsubseteq D_1, \ldots, C_n \sqsubseteq D_n\}$ into the concept satisfiability problem in $\mathcal{ALC}_{reg}$ (w.r.t. the empty TBox). Since the problem of concept satisfiability w.r.t. a TBox is trivially harder than checking the satisfiability of a TBox, that paper extends the result given in (Baader et al., 1990).

The technique exploited in (Baader et al., 1990) and (Baader, 1990a) is based on the notion of *concept tree*. A concept tree is generated starting from a concept $C$ in order to check its satisfiability (or universal satisfiability). The way a concept tree is generated from a concept $C$ is similar in flavor to the way a complete constraint system is generated from the constraint system $\{x : C\}$. However, the extension of the concept tree method to deal with number restrictions and individuals in the knowledge base is neither obvious, nor suggested in the cited papers; on the other hand, the extension of the calculus based on constraint systems is immediate, provided that additional features have a counterpart in First Order Logic.

In (Schild, 1991) some results more general than those in (Baader, 1990a) are obtained by considering languages more expressive than $\mathcal{ALC}_{reg}$ and dealing with the concept satisfiability problem in such languages. The results are obtained by establishing a correspondence between concept languages and Propositional Dynamic Logics (PDL), and reducing the given problem to a satisfiability problem in PDL. Such an approach allows Schild to find several new results exploiting known results in the PDL framework. However, it cannot be used to deal with every concept language. In fact, the correspondence cannot be established when the language includes some concept constructors having no counterpart in PDL (e.g., number restrictions, or individuals in an ABox).

Recently, an algebraic approach to cycles has been proposed in (Dionne et al., 1992), in which (possibly cyclic) definitions are interpreted as determining an equivalence relation over the terms describing concepts. The existence and uniqueness of such an equivalence relation derives from Aczel's results on non-well founded sets. In (Dionne et al., 1993) the same researchers prove that subsumption based on this approach is equivalent to subsumption in greatest fixpoint semantics. The language analyzed is a small fragment of the one used in the TKRS K-REP, and contains conjunction and existential-universal quantifications combined into one construct (hence it is similar to $\mathcal{FL}_0$). The difficulty of extending these results lies in the fact that it is not clear how individuals can be interpreted in this algebraic setting. Moreover, we believe that constructive approaches like the algebraic one, give counterintuitive results when applied to non-constructive features of concept languages—as negation and number restrictions.

In conclusion, all these approaches, i.e., reduction to automata problems, concept trees, reduction to PDL and algebraic semantics, deal only with TBoxes and they don't seem to be suitable to deal also with ABoxes. On the other hand, the constraint system technique, even though it was conceived for TBox-reasoning, can be easily extended to ABox-reasoning, as also shown in (Hollunder, 1990; Baader & Hollunder, 1991; Donini et al., 1993).

## 4.2 Inclusions versus Concept Definitions

Now we compare the expressive power of TBoxes defined as a set of inclusions (as done in this paper) and TBoxes defined as a set of (possibly cyclic) concept introductions of the form $A \stackrel{.}{\sqsubseteq} D$ and $A \doteq D$.

Unlike (Baader, 1990a) and (Schild, 1991), we consider reasoning problems dealing with TBox and ABox together. Moreover, we use the descriptive semantics for the concept introductions, as we do for inclusions. The result we have obtained is that inclusion statements and concept introductions actually have the same expressive power. In detail, we show that the satisfiability of a knowledge base $\Sigma = \langle \mathcal{A}, \mathcal{T} \rangle$, where $\mathcal{T}$ is a set of inclusion statements, can be reduced to the satisfiability of a knowledge base $\Sigma' = \langle \mathcal{A}', \mathcal{T}' \rangle$ such that $\mathcal{T}'$ is a set of concept introductions. The other direction, from concept introductions to inclusions, is trivial since introductions of the form $A \doteq D$ can be expressed by the pair of inclusions $A \sqsubseteq D$ and $D \sqsubseteq A$, while a concept name specification $A \stackrel{.}{\sqsubseteq} D$ can be rewritten as the inclusion $A \sqsubseteq D$ (as already mentioned in Section 2).

As a notation, given a TBox $\mathcal{T} = \{C_1 \sqsubseteq D_1, \ldots, C_n \sqsubseteq D_n\}$, we define the concept $C_\mathcal{T}$ as $C_\mathcal{T} = (\neg C_1 \sqcup D_1) \sqcap \cdots \sqcap (\neg C_n \sqcup D_n)$. As pointed out in (Baader, 1990a) for $\mathcal{ALC}$, an interpretation satisfies a TBox $\mathcal{T}$ if and only if it satisfies the equation $C_\mathcal{T} = \top$. This result easily extends to $\mathcal{ALCNR}$, as stated in the following proposition.

**Proposition 4.1** *Given an $\mathcal{ALCNR}$-TBox $\mathcal{T} = \{C_1 \sqsubseteq D_1, \ldots, C_n \sqsubseteq D_n\}$, an interpretation $\mathcal{I}$ satisfies $\mathcal{T}$ if and only if it satisfies the equation $C_\mathcal{T} = \top$.*

**Proof.** An interpretation $\mathcal{I}$ satisfies an inclusion $C \sqsubseteq D$ if and only if it satisfies the equation $\neg C \sqcup D = \top$; $\mathcal{I}$ satisfies the set of equations $\neg C_1 \sqcup D_1 = \top, \ldots, \neg C_n \sqcup D_n = \top$ if and only if $\mathcal{I}$ satisfies $(\neg C_1 \sqcup D_1) \sqcap \cdots \sqcap (\neg C_n \sqcup D_n) = \top$. The claim follows. $\qquad\square$

Given a knowledge base $\Sigma = \langle \mathcal{A}, \mathcal{T} \rangle$ and a concept $A$ not appearing in $\Sigma$, we define the knowledge base $\Sigma' = \langle \mathcal{A}', \mathcal{T}' \rangle$ as follows:

$$\mathcal{A}' = \mathcal{A} \cup \{A(b) \mid b \text{ is an individual in } \Sigma\}$$
$$\mathcal{T}' = \{A \doteq C_{\mathcal{T}} \sqcap \forall P_1.A \sqcap \cdots \sqcap \forall P_n.A\}$$

where $P_1, P_2, \ldots, P_n$ are all the role names appearing in $\Sigma$. Note that $\mathcal{T}'$ has a single inclusion, which could be also thought of as one primitive concept specification.

**Theorem 4.2** $\Sigma = \langle \mathcal{A}, \mathcal{T} \rangle$ *is satisfiable if and only if* $\Sigma' = \langle \mathcal{A}', \mathcal{T}' \rangle$ *is satisfiable.*

*Proof.* In order to simplify the machinery of the proof, we will use for $\mathcal{T}'$ the following (logically equivalent) form:

$$\mathcal{T}' = \{A \sqsubseteq C_{\mathcal{T}}, A \sqsubseteq \forall P_1.A, \ldots, A \sqsubseteq \forall P_n.A\}$$

(Note that we use the symbol '$\sqsubseteq$' instead of '$\doteq$' because now the concept name $A$ appears as the left-hand side of many statements, we must consider these statements as inclusions).

"$\Rightarrow$" Suppose $\Sigma = \langle \mathcal{A}, \mathcal{T} \rangle$ satisfiable. From Theorem 3.7, there exists a complete constraint system $S$ without clash, which defines a canonical interpretation $\mathcal{I}_S$ which is a model of $\Sigma$. Define the constraint system $S'$ as follows:

$$S' = S \cup \{w : A \mid w \text{ is an object in } S\}$$

and call $\mathcal{I}_{S'}$ the canonical interpretation associated to $S'$. We prove that $\mathcal{I}_{S'}$ is a model of $\Sigma'$.

First observe that every assertion in $\mathcal{A}$ is satisfied by $\mathcal{I}_{S'}$ since $\mathcal{I}_{S'}$ is equal to $\mathcal{I}_S$ except for the interpretation of $A$, and $A$ does not appear in $\mathcal{A}$. Therefore, every assertion in $\mathcal{A}'$ is also satisfied by $\mathcal{I}_{S'}$, either because it is an assertion of $\mathcal{A}$, or (if it is an assertion of the form $A(b)$) by definition of $S'$.

Regarding $\mathcal{T}'$, note that by definition of $S'$, we have $A^{\mathcal{I}_{S'}} = \Delta^{\mathcal{I}_{S'}} = \Delta^{\mathcal{I}_S}$; therefore both sides of the inclusions of the form $A \sqsubseteq \forall P_i.A$ $(i = 1, \ldots, n)$ are interpreted as $\Delta^{\mathcal{I}_{S'}}$, hence they are satisfied by $\mathcal{I}_{S'}$. Since $A$ does not appear in $C_{\mathcal{T}}$, we have that $(C_{\mathcal{T}})^{\mathcal{I}_{S'}} = (C_{\mathcal{T}})^{\mathcal{I}_S}$. Moreover, since $\mathcal{I}_S$ satisfies $\mathcal{T}$, we also have, by Proposition 4.1, that $(C_{\mathcal{T}})^{\mathcal{I}_S} = \Delta^{\mathcal{I}_S}$, therefore $(C_{\mathcal{T}})^{\mathcal{I}_{S'}} = (C_{\mathcal{T}})^{\mathcal{I}_S} = \Delta^{\mathcal{I}_S} = \Delta^{\mathcal{I}_{S'}}$. It follows that also both sides of the inclusion $A \sqsubseteq C_{\mathcal{T}}$ are interpreted as $\Delta^{\mathcal{I}_{S'}}$. In conclusion, $\mathcal{I}_{S'}$ satisfies $\mathcal{T}'$.

"$\Leftarrow$" Suppose $\Sigma' = \langle \mathcal{A}', \mathcal{T}' \rangle$ satisfiable. Again, because of Theorem 3.7, there exists a complete constraint system $S'$ without clash, which defines a canonical interpretation $\mathcal{I}_{S'}$ which is a model of $\Sigma'$. We show that $\mathcal{I}_{S'}$ is also a model of $\Sigma$.

First of all, the assertions in $\mathcal{A}$ are satisfied because $\mathcal{A} \subseteq \mathcal{A}'$, and $\mathcal{I}_{S'}$ satisfies every assertion in $\mathcal{A}'$. To prove that $\mathcal{I}_{S'}$ satisfies $\mathcal{T}$, we first prove the following equation:

$$A^{\mathcal{I}_{S'}} = \Delta^{\mathcal{I}_{S'}} \tag{2}$$

Equation 2 is proved by showing that, for every object $s \in \Delta^{\mathcal{I}_{S'}}$, $s$ is in $A^{\mathcal{I}_{S'}}$. In order to do that, observe a general property of constraint systems: Every variable in $S'$ is a successor of an individual. This comes from the definition of the generating rules, which add variables to the constraint system only as direct successors of existing objects, and at the beginning $S_{\Sigma'}$ contains only individuals.

Then, Equation 2 is proved by observing the following three facts:

1. for every individual $b$ in $\Delta^{\mathcal{I}_{S'}}$, $b \in A^{\mathcal{I}_{S'}}$;

2. if an object $s$ is in $A^{\mathcal{I}_{S'}}$, then because $\mathcal{I}_{S'}$ satisfies the inclusions $A^{\mathcal{I}_{S'}} \subseteq (\forall P_1.A)^{\mathcal{I}_{S'}}, \ldots,$ $A^{\mathcal{I}_{S'}} \subseteq (\forall P_n.A)^{\mathcal{I}_{S'}}$, every direct successor of $s$ is in $A^{\mathcal{I}_{S'}}$;

3. the successor relation is closed under the direct successor relation

From the Fundamental Theorem on Induction (see e.g., Wand, 1980, page 41) we conclude that every object $s$ of $\Delta^{\mathcal{I}_{S'}}$ is in $A^{\mathcal{I}_{S'}}$. This proves that Equation 2 holds.

From Equation 2, and the fact that $\mathcal{I}_{S'}$ satisfies the inclusion $A^{\mathcal{I}_{S'}} \subseteq (C_{\mathcal{T}})^{\mathcal{I}_{S'}}$, we derive that $(C_{\mathcal{T}})^{\mathcal{I}_{S'}} = \Delta^{\mathcal{I}_{S'}}$, that is $\mathcal{I}_{S'}$ satisfies the equation $C_{\mathcal{T}} = \top$. Hence, from Proposition 4.1, $\mathcal{I}_{S'}$ satisfies $\mathcal{T}$, and this completes the proof of the theorem. ☐

The machinery present in this proof is not new. In fact, realizing that the inclusions $A \sqsubseteq \forall P_1.A, \ldots, A \sqsubseteq \forall P_n.A$ simulate a transitive closure on the roles $P_1, \ldots, P_n$, one can recognize similarities with the proofs given by Schild (1991) and Baader (1990a). The difference is that their proofs rely on the notion of *connected model* (Baader uses the equivalent notion of *rooted* model). In contrast, the models we obtain are not connected, when the individuals in the knowledge base are not. What we exploit is the weaker property that every variable in the model is a successor of an individual.

Note that the above reduction strongly relies on the fact that disjunction '⊔' and complement '¬' are within the language. In fact, disjunction and complement are necessary in order to express all the inclusions of a TBox $\mathcal{T}$ inside the concept $C_{\mathcal{T}}$. Therefore, the proof holds for $\mathcal{ALC}$-knowledge bases, but does not hold for TKRSs not allowing for these constructors of concepts (e.g., BACK).

Furthermore, for the language $\mathcal{FL}_0$ introduced in Section 4.1, the opposite result holds. In fact, McAllester (1991) proves that computing subsumption w.r.t. a set of inclusions is EXPTIME-hard, even in the small language $\mathcal{FL}_0$. Conversely, Nebel (1991) proves that subsumption w.r.t. a set of cyclic definitions in $\mathcal{FL}_0$ can be done in PSPACE. Combining the two results, we can conclude that for $\mathcal{FL}_0$ subsumption w.r.t. a set of inclusions and subsumption w.r.t. a set of definitions are in different complexity classes, hence (assuming EXPTIME $\neq$ PSPACE) inclusion statements are strictly more expressive than concept definitions in $\mathcal{FL}_0$.

It is still open whether inclusions and definitions are equivalent in languages whose expressivity is between $\mathcal{FL}_0$ and $\mathcal{ALC}$.

## 5. Discussion

In this paper we have proved the decidability of the main inference services of a TKRS based on the concept language $\mathcal{ALCNR}$. We believe that this result is not only of theoretical importance, but bears some impact on existing TKRSs, because a complete procedure can be easily devised from the calculus provided in Section 3. From this procedure, one can build more efficient (but still complete) ones, as described at the end of Section 3.2, and also by applying standard optimization techniques such as those described in (Baader, Hollunder, Nebel, Profitlich, & Franconi, 1992). An optimized procedure can perform well for small sublanguages where reasoning is tractable, while still being complete when solving more complex tasks. However, such a complete procedure will still take exponential time and space in the worst case, and it may be argued what could be its practical applicability. We comment in following on this point.

Firstly, a complete procedure (possibly optimized) offers a benchmark for comparing incomplete procedures, not only in terms of performance, but also in terms of missed inferences. Let us illustrate this point in detail, by providing a blatant paradox: consider the mostly incomplete constant-time procedure, answering always "No" to any check. Obviously this useless procedure outperforms any other one, if missed inferences are not taken into account. This paradox shows that incomplete procedures can be meaningfully compared only if missed inferences are considered. But to recognize missed inferences over large examples, one needs exactly a complete procedure—even if not an efficient one—like ours. We believe that a fair detection of missed inferences would be of great help even when the satisfaction of end users is the primary criterion for judging incomplete procedures.

Secondly, a complete procedure can be used for "anytime classification", as proposed in (MacGregor, 1992). The idea is to use a fast, but incomplete algorithm as a first step in analyzing the input knowledge, and then do more reasoning in background. In the cited paper, resolution-based theorem provers are proposed for performing this background reasoning. We argue that any specialized complete procedure will perform better than a general theorem prover. For instance, theorem provers are usually not specifically designed to deal with filtration techniques.

Moreover, our calculus can be easily adapted to deal with rules. As outlined in the introduction, rules are often used in practical TKRSs. Rules behave like one-way concept inclusions—no contrapositive is allowed—and they are applied only to known individuals. Our result shows that rules in $\mathcal{ALCNR}$ can be applied also to unknown individuals (our variables in a constraint system) without endangering decidability. This result is to be compared with the negative result in (Baader & Hollunder, 1992), where it is shown that subsumption becomes undecidable if rules are applied to unknown individuals in CLASSIC.

Finally, the calculus provides a new way of building incomplete procedures, by modifying some of the propagation rules. Since the rules build up a model, modifications to them have a semantical counterpart which gives a precise account of the incomplete procedures obtained. For example, one could limit the size of the canonical model by a polynomial in the size of the KB. Semantically, this would mean to consider only "small" models, which is reasonable when the intended models for the KB are not much bigger than the size of the KB itself. We believe that this way of designing incomplete procedures "from above", i.e., starting with the complete set of inferences and weakening it, is dual to the way incomplete procedures have been realized so far "from below", i.e., starting with already incomplete inferences and adding inference power by need.

Further research is still needed to address problems issuing from practical systems. For example, to completely express role restrictions inside number restrictions, qualified number restrictions (Hollunder & Baader, 1991) should be taken into account. Also, the language resulting from the addition of enumerated sets (called ONE-OF in CLASSIC), and role fillers to $\mathcal{ALCNR}$ is still to be studied, although it does not seem to endanger the filtration method we used. Instead, a different method might be necessary if inverse roles are added to $\mathcal{ALCNR}$, since the finite model property is lost (as shown in Schild, 1991). Finally, the addition of concrete domains (Baader & Hanschke, 1991) remains open.

## Acknowledgements

We thank Maurizio Lenzerini for the inspiration of this work, as well as for several discussions that contributed to the paper. Werner Nutt pointed out to us the observation mentioned at the end of Section 3, and we thank him and Franz Baader for helpful comments on earlier drafts. We thank also the anonymous reviewers, whose stimulating comments helped us in improving on the submitted version.

The research was partly done while the first author was visiting the Dipartimento di Informatica e Sistemistica, Università di Roma "La Sapienza". The third author also acknowledges Yoav Shoham for his hospitality at the Computer Science Department of Stanford University, while the author was developing part of this research.

This work has been supported by the ESPRIT Basic Research Action N.6810 (COMPULOG 2) and by the Progetto Finalizzato Sistemi Informatici e Calcolo Parallelo of the CNR (Italian Research Council), LdR "Ibridi".